%% file: bare_conf.tex
\documentclass[conference, a4paper]{IEEEtran}

\ifCLASSINFOpdf
 
\else
 
\fi

\usepackage{graphicx}
\graphicspath{{./}}
\usepackage{filecontents}
\usepackage{url}
\usepackage[bookmarks=false]{hyperref}

\usepackage{amsmath}
\usepackage{multirow}
\begin{document}

\title{Effect of Improper Segmentation on Word Spotting} 

\author{\IEEEauthorblockN{
Sounak Dey\IEEEauthorrefmark{1},
Anguelos Nicolaou\IEEEauthorrefmark{1},
Josep Llados\IEEEauthorrefmark{1}, and
Umapada Pal\IEEEauthorrefmark{2}
}
\IEEEauthorblockA{
\IEEEauthorrefmark{1}
Computer Vision Center, Edifici O, Universitat Autonoma de Barcelona,Bellaterra, Spain
\\ 
\IEEEauthorrefmark{2}
CVPR Unit, Indian Statistical Institute, India  
\\
Email: \href{mailto:sdey@cvc.uab.es}{sdey@cvc.uab.es},
\href{mailto:anguelos@cvc.uab.es}{anguelos@cvc.uab.es}, 
\href{mailto:josep@cvc.uab.es}{josep@cvc.uab.es},
\href{mailto:umapada@isical.ac.in}{umapada@isical.ac.in}
}

}

\maketitle

\input{ch_abstract.tex}

\IEEEpeerreviewmaketitle

\input{tbl_qualitative.tex}
\input{ch_introduction.tex}

\input{ch_stateOfArt.tex}

\input{ch_experiments.tex}

\input{ch_conclusion.tex}

\section*{Acknowledgment}
The authors would like to thank Marcal Rusi\~nol and Suman Ghosh for running the BoVW and Fischer CCA experiments as well as David Fern\`andez for providing a tuned implementation of~\cite{manmatha2005scale}.

\bibliographystyle{IEEEtran}
\bibliography{references.bib}

\end{document}

%% file: ch_abstract.tex
\begin{abstract}
Word spotting is an important recognition task in historical document analysis.
In most cases methods are developed and evaluated assuming perfect word segmentations.
In this paper we propose an experimental framework to quantify the effect of goodness of word segmentation has on the performance achieved by word spotting methods in identical unbiased conditions.
The framework consists of generating systematic distortions on segmentation and retrieving the original queries from the distorted dataset.
We apply the framework on the George Washington and Barcelona Marriage Dataset and on several established and state-of-the-art methods.
The experiments allow for an estimate of the end-to-end performance of word spotting methods. 
\end{abstract}

%% file: tbl_qualitative.tex
\begin{figure*}
\begin{tabular}{ll}
\begin{tabular}{p{.06\textwidth}}Query sample\end{tabular} & \begin{tabular}{l}\includegraphics[width=.14\textwidth]{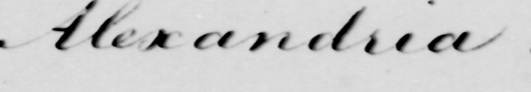} \end{tabular} \\ 
\begin{tabular}{p{.07\textwidth}}Random retrieval\end{tabular} & \begin{tabular}{c}\includegraphics[width=.07\textwidth]{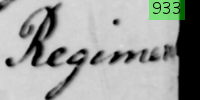} \includegraphics[width=.07\textwidth]{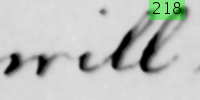} \includegraphics[width=.08\textwidth]{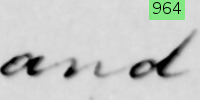} \includegraphics[width=.08\textwidth]{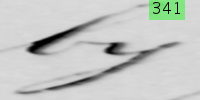} \includegraphics[width=.08\textwidth]{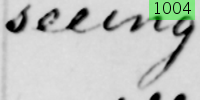} \includegraphics[width=.08\textwidth]{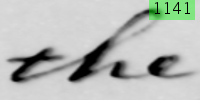} \includegraphics[width=.08\textwidth]{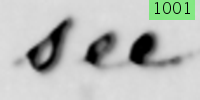} \includegraphics[width=.08\textwidth]{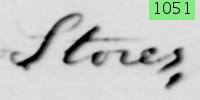} \includegraphics[width=.08\textwidth]{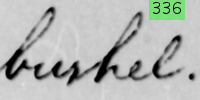} \includegraphics[width=.08\textwidth]{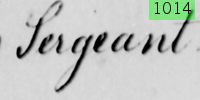}\end{tabular} \\ 
 \begin{tabular}{p{.06\textwidth}}DTW\end{tabular}   &  \begin{tabular}{c} \includegraphics[width=.08\textwidth]{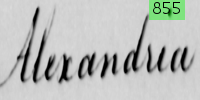} \includegraphics[width=.08\textwidth]{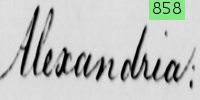} \includegraphics[width=.08\textwidth]{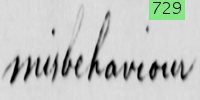} \includegraphics[width=.08\textwidth]{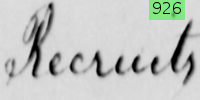} \includegraphics[width=.08\textwidth]{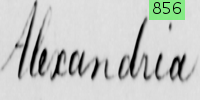} \includegraphics[width=.08\textwidth]{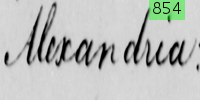} \includegraphics[width=.08\textwidth]{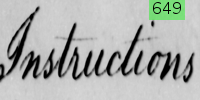} \includegraphics[width=.08\textwidth]{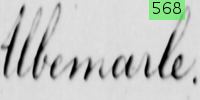} \includegraphics[width=.08\textwidth]{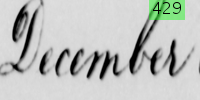} \includegraphics[width=.08\textwidth]{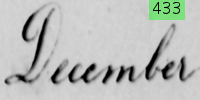}\end{tabular} \\ 
 \begin{tabular}{p{.06\textwidth}}Quad-Tree\end{tabular}   &  \begin{tabular}{c} \includegraphics[width=.08\textwidth]{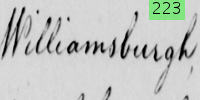} \includegraphics[width=.08\textwidth]{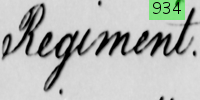} \includegraphics[width=.08\textwidth]{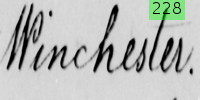} \includegraphics[width=.08\textwidth]{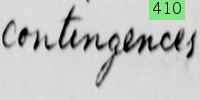} \includegraphics[width=.08\textwidth]{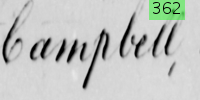} \includegraphics[width=.08\textwidth]{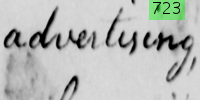} \includegraphics[width=.08\textwidth]{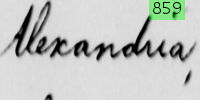} \includegraphics[width=.08\textwidth]{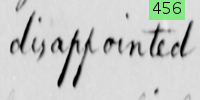} \includegraphics[width=.08\textwidth]{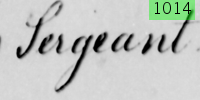} \includegraphics[width=.08\textwidth]{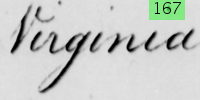}\end{tabular} \\ 
 \begin{tabular}{p{.06\textwidth}}HOG\end{tabular}   &  \begin{tabular}{c} \includegraphics[width=.08\textwidth]{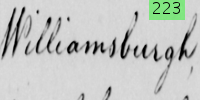} \includegraphics[width=.08\textwidth]{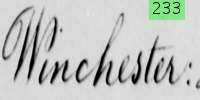} \includegraphics[width=.08\textwidth]{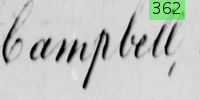} \includegraphics[width=.08\textwidth]{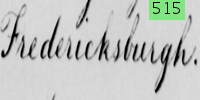} \includegraphics[width=.08\textwidth]{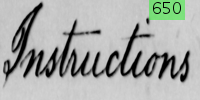} \includegraphics[width=.08\textwidth]{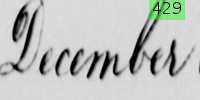} \includegraphics[width=.08\textwidth]{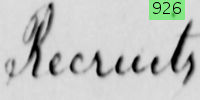} \includegraphics[width=.08\textwidth]{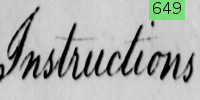} \includegraphics[width=.08\textwidth]{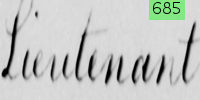} \includegraphics[width=.08\textwidth]{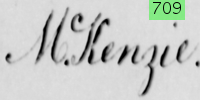}\end{tabular} \\ 
 \begin{tabular}{p{.06\textwidth}}LBP\end{tabular}   &  \begin{tabular}{c} \includegraphics[width=.08\textwidth]{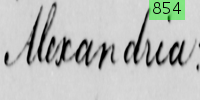} \includegraphics[width=.08\textwidth]{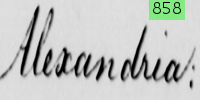} \includegraphics[width=.08\textwidth]{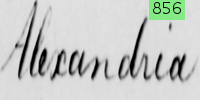} \includegraphics[width=.08\textwidth]{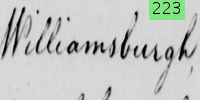} \includegraphics[width=.08\textwidth]{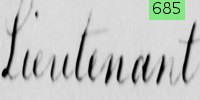} \includegraphics[width=.08\textwidth]{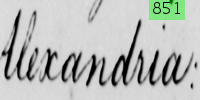} \includegraphics[width=.08\textwidth]{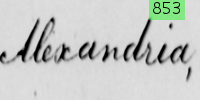} \includegraphics[width=.08\textwidth]{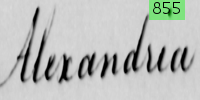} \includegraphics[width=.08\textwidth]{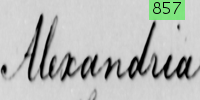} \includegraphics[width=.08\textwidth]{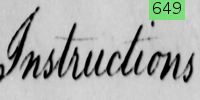}\end{tabular} \\ 
 \begin{tabular}{p{.05\textwidth}}BoVW\end{tabular}   &  \begin{tabular}{c} \includegraphics[width=.08\textwidth]{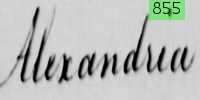} \includegraphics[width=.08\textwidth]{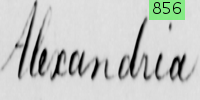} \includegraphics[width=.08\textwidth]{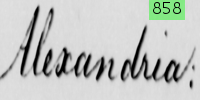} \includegraphics[width=.08\textwidth]{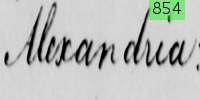} \includegraphics[width=.08\textwidth]{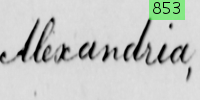} \includegraphics[width=.08\textwidth]{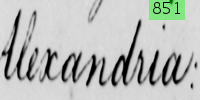} \includegraphics[width=.08\textwidth]{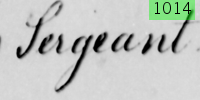} \includegraphics[width=.08\textwidth]{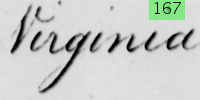} \includegraphics[width=.08\textwidth]{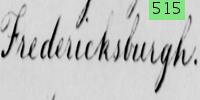} \includegraphics[width=.08\textwidth]{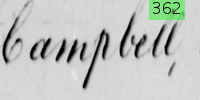}\end{tabular} \\ 
 \begin{tabular}{p{.05\textwidth}}FV CCA\end{tabular}   &  \begin{tabular}{c} \includegraphics[width=.08\textwidth]{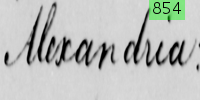} \includegraphics[width=.08\textwidth]{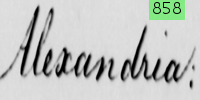} \includegraphics[width=.08\textwidth]{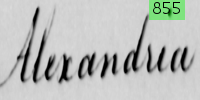} \includegraphics[width=.08\textwidth]{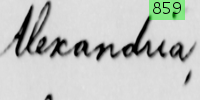} \includegraphics[width=.08\textwidth]{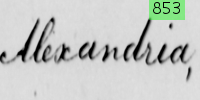} \includegraphics[width=.08\textwidth]{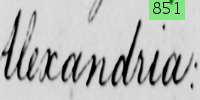} \includegraphics[width=.08\textwidth]{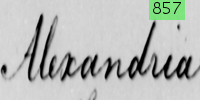} \includegraphics[width=.08\textwidth]{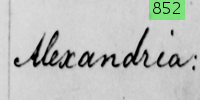} \includegraphics[width=.08\textwidth]{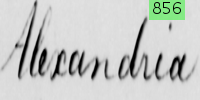} \includegraphics[width=.08\textwidth]{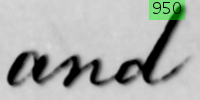}\end{tabular}

\end{tabular}
\label{fig:retrieval_example}
\caption{Randomly selected qualitative example of retrievals for all methods. From left to right, are the nearest samples retrieved by each method. The id of each sample which distinguishes different samples with the same text is marked superscript in green.}
\end{figure*}

%% file: ch_introduction.tex
\section{Introduction}
In recent years, many important and valuable documents are accessible only as imaged format.
A larger amount of documents which were previously in paper, is also being digitised as images.
These documents are quite precious and important for humanity and it have to be  preserved permanently.
These images contain printed or more often handwritten text. 
Most of these documents have been degraded to a great extent due to reading and ageing processes. 
These  documents should be provided for users to access and retrieve including searching keywords through out the documents.
Word spotting has been proposed as an alternative to OCR, as a form of content-based retrieval procedure, which results in a ranked list of word images that are similar to the query word. 
The query can be either an example image Query by Example (QBE)\cite{rusinol2015efficient} or an string containing the word to be searched Query by String (QBS)\cite{almazan2012efficient}.
The basic idea of word spotting using query by example is that a template image is selected form a set of predefined keywords, i.e. words of interest and then search is initiated to find out its other instances in the target set of the digitized documents.
This factoid makes the approach more flexible and suitable for indexing and retrieval of degraded and historical documents written in multiple languages.
In a typical word spotting pipeline, pre-processed patches/imperfectly segmented or cropped words are first obtained.
This segmentation step is not always straightforward and might be prone to many errors. 
In fact, although word and text line segmentation is a highly cultured research topic, it is far from being a solved problem.
In the literature, word spotting evolves under two distinct sections: the segmentation-free approach and the segmentation-based approach.
Most of the segmentation free word spotting methods uses a seperate segmentation technique adjoint to the word spotting method.
The proposed word candidates by different techiniques is of prime importance for the word spotting algorithm to work.
Depending on the amount of information localised, the performance of the word spotting method differs a lot.
In a realistic scenario getting a perfectly segmented word from the database is a very rare phenomenon.
On the other hand, the performance of the good state-of-art (SOA) methods degrades significantly if words are improperly segmented.
The main motivation of this work is to provide an exhaustive analysis of different SOA methods in a practical scenario.
This allows to define a taxonomy on the convenience of each methos to the possibility of proper segmentation in target documents.
Thus, a "difficult" document would require a spotting method that is robust.
This analysis can bridge the gap between both the segmentation based and segmentation free word spotting methods.
Segmentation errors have a cumulative effect on subsequent word representations and matching steps. 
This dependence on good word segmentation motivated the researchers from the keyword-spotting domain to recently move towards complete segmentation-free methods\cite{rusinol2015efficient}.
But most of these end to end word spotting method comprise a segmenter whose quality of segmentation is not always perfect.
In this paper we analyse the robustness of different state of the art method to improper segmentation.
We provide mean average precision measures of different levels of cropping.
Additionally we provide some other measures too.
Cross domain/dataset performance by different methods.
We also present the orthogonality/independence of different such methods.
The final goal of the provided evaluation is to have a recommensation survey not only on the robustness of the methods in terms of quality segmentations, but on their complementarity in fusion techniques.

The rest of this paper is structured as follows. In section \ref{soa} we give an overview of the state of the art. Then we describe the experimental framework for analysis and results \ref{experiments}. Finally in section \ref{conclusions} we conclude with the contribution and ideas about future work regarding this framework.

%% file: ch_stateOfArt.tex
\section{State of the art}
\label{soa}
Word spotting can be broadly classified under two distinct sections: the segmentation-free approach and the segmentation-based approach. In the later approach, there is a tremendous effort towards solving the word segmentation problem\cite{rath2003features}
 \cite{bhardwaj2008script}.
 One of the main challenges of keyword spotting methods, either learning-free or learning-based, is that they usually need to segment the document images into words \cite{howe2005boosted}
 \cite{liang2012synthesised} or text lines \cite{frinken2012novel}
 using a layout analysis step.  
 In critical scenarios dealing with handwritten text and highly degraded documents \cite{likforman2007text}\cite{louloudis2009text} segmentation is highly crucial.  
 The work of Rusiñol et al.\cite{rusinol2011browsing} avoids segmentation by representing regions with a fixed-length descriptor based on the well-known bag of visual words (BoW) framework \cite{csurka2004visual}. 
 In this case, comparison of regions is much faster with the use of a dot product or Euclidean distance. 
 Late works on word spotting have proposed methods where a precise word segmentation is not required, or, in some cases, no segmentation at all. 
 The recent works of Rodriguez et.al. \cite{rodriguez2012model} propose methods that relax the segmentation problem by requiring only segmentation at the text line level. 
 In \cite{gatos2009segmentation}, Gatos and Pratikakis perform a fast and very coarse segmentation of the page to detect salient text regions.

    The most common approach to use a patch- based framework in which a window slides over the whole document.
     In such a framework expected segmentations may not be perfect and elements from surrounding words will appear within a patch. Automatic word segmentation, as presented in \cite{srihari2006spotting}, is based on taking several features on either side of a potential segmentation point and then using a neural network for deciding whether or not the segmentation is between two distinct words.
  The segmentation free method attempts to perform spotting and segmentation concurrently. 
  An entire line image acts as input in place of a candidate word image.
 The text detection algorithms can be broadly classified into two categories: connected component based approach and sliding window-based approaches\cite{lee2011adaboost}\cite{wang2011end}.
   The main drawback is that the number of rectangles that it needs to be assessed grows rapidly when text with different scale, aspect, rotation and other variations are taken into consideration. 

In the same spirit with the aforementioned approaches, this paper concerns a study on the the amount of overlapping that is needed with respect to the actual ground-truth for recreating the accuracy at first place achieved with different pipeline in the case of segmented words.
\section{Methods compared}
\subsubsection{\textbf{Almazan et.al. \cite{almazan2012efficient}}}
 The use of exemplar SVM's has created one of the best segmentation free methods in the literature based on accuracy and mean average precision.
 This supervised method represents the documents with a grid of Histogram of Gradients (HOG) descriptors. An exemplar SVM framework is used to produce a better representation of the query.
They also uses a more discriminative representation based on Fisher Vector to rerank the best regions retrieved, which in turn is used to expand the Exempler SVM training set and improve the query representation.
Ultimately the document descriptor is pre-calculated and compressed with the Product Quantanization.
\subsubsection{\textbf{Rusi\~nol et.al \cite{rusinol2015efficient}}}
In the paradigm of segmentation free, query by example handwritten word spotting, it has out performed the recent state of the art keyword spotting approaches.
In the experiments of this paper for the state of comparision in the proposed framework, we have used a segmentation based variant.
This method uses a patch based framework where local patches which are specified by bag of visual words model powered by SIFT descriptors.
This descriptors are then projected to topic space with the Latent Semantic Analysis technique and then compressing of these descriptors is done with the Product Quantization method.
This statistical approach is in turn enables a efficient indexation of document information both in terms of memory and time. 

\subsubsection{\textbf{Rath et.al \cite{rath2003word}}}
It can be considered the baseline algorithm for matching handwritten words in noisy historical documents.
The segmented word images are preprocessed to create sets of 1-dimensional features, which are then compared using Dynamic Time Warping (DTW).

\subsubsection{\textbf{Method based on Quad Tree}}
This method relises on a adaptive feature extraction technique \cite{vamvakas2010handwritten} based on recursive subdivisions of the word images so that the resulting sub images at each iteration have balanced (approximately equal) numbers of foreground pixels, for two levels.
This adaptive hierarchical decomposition technique which determines the pyramidal grid that is recursively updated through the calculation of image geometric centroids \cite{sidiropoulos2011content}. 

\subsubsection{\textbf{Method based on Local Binary Pattern}}
In this method the apative hierarchical decomposition technique employs the pooling of the Local Binary Patterns in the adaptive regions, which are determined by a pyramidal grid that is recursively updated through the calculation of image geometric centroids to calculate the feature vector for matching.\footnote{Note to the reviewers: This is a paper under review and available on arxiv under the title\textit{Local Binary Pattern for word spoting in handwritten historical documents.} A proper bibliographic reference will be added.}

\subsubsection{\textbf{Method based on HOG}}
Similar to the previous method feature vector is created by pooling the gradients in the similar pyramidal grid.

\subsection{Fusion}
Since we provide an analysis of complementarity, a naive late fusion between methods is also proposed.
Due to the fact that many of the methods analysed in this paper are learning-free, a learning-free late fusion was preferred.
More specifically the fundamental assumption is that any method that does retrieval, can provide a vector with all distances between the query sample and all samples in the retrieval database.
The fusion method we propose is a weighted sum of such vectors two or more methods provide.
In the taxonomy of fusion methods described in~\cite{atrey2010multimodal}, it would be described as a weighted linear fusion at the level of decision.
One of the main benefits of this method is that it can be applied to feature representations of a variable size or even methods that don't have a feature representation such as DTW.

%% file: ch_experiments.tex
\begin{figure*}[!t]
\begin{minipage}[b]{0.24\linewidth}
    \includegraphics[width=\textwidth]{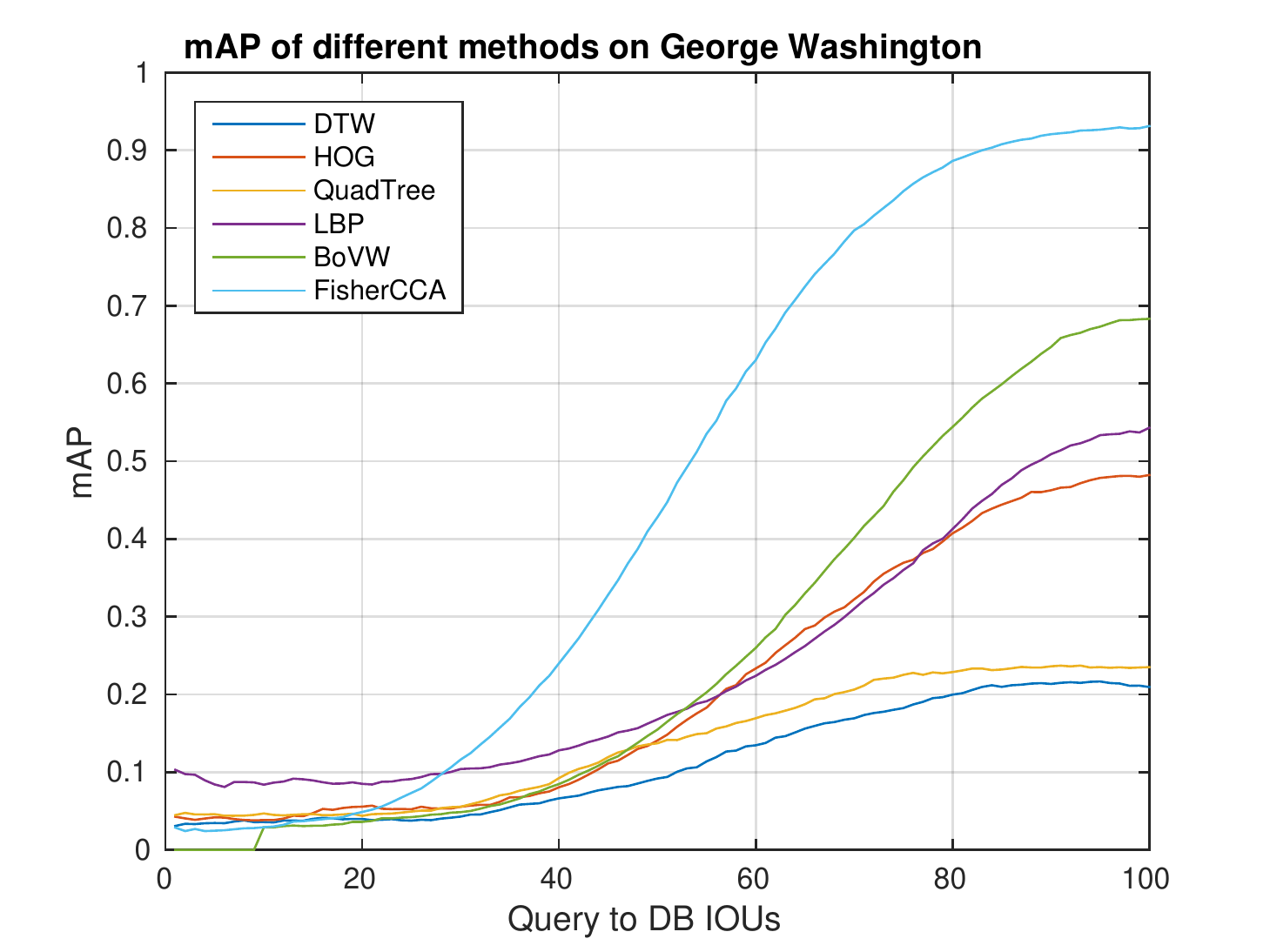}
    \caption{Effect of improper segmentation on mAP for GW.}
  	\label{fig:mAP_GW}
\end{minipage}
\begin{minipage}[b]{0.24\linewidth}
    \includegraphics[width=\textwidth]{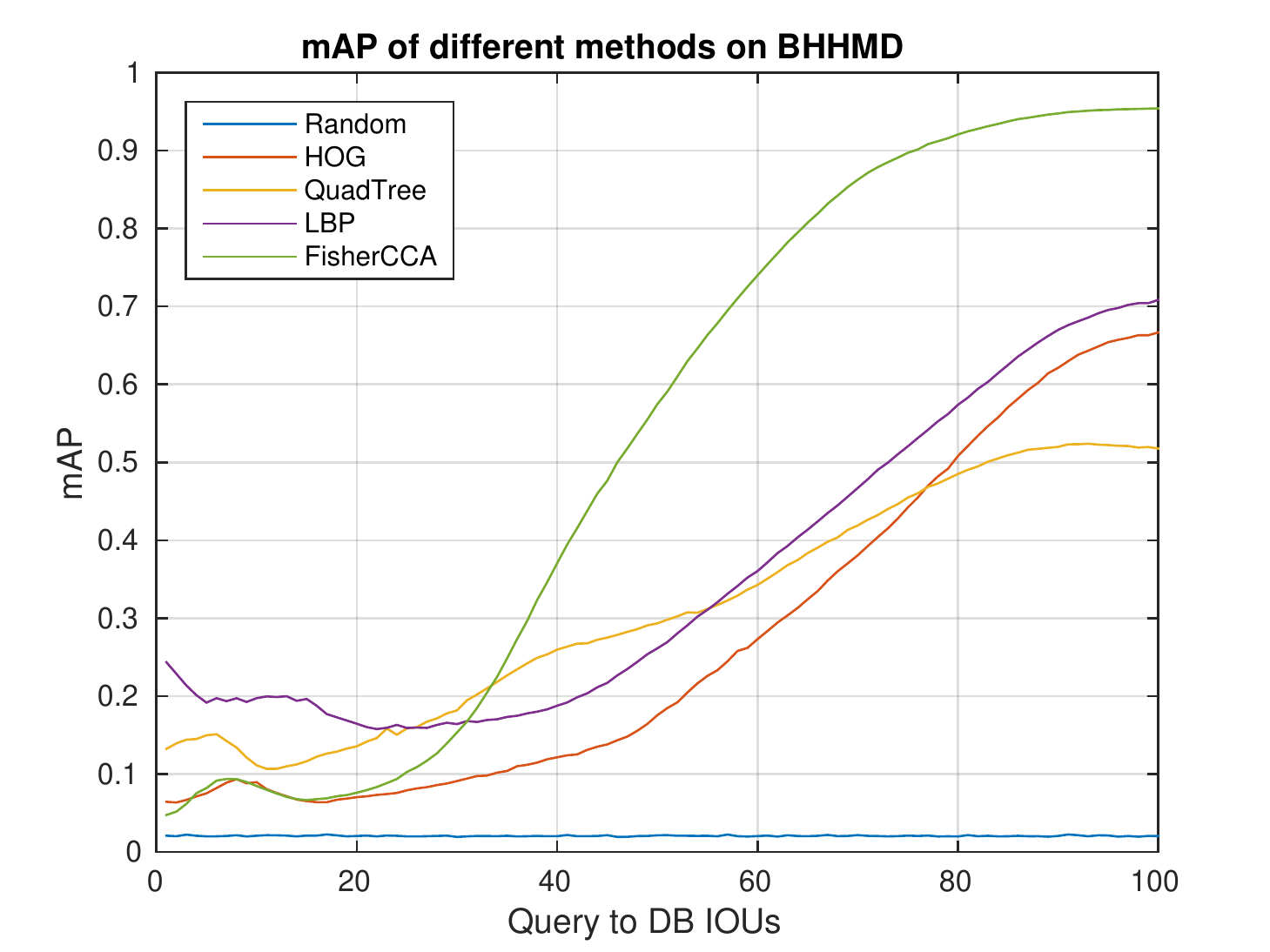}
    \caption{Effect of improper segmentation on mAP for BHHWD.}
	\label{fig:mAP_BCN}
\end{minipage}
\begin{minipage}[b]{0.24\linewidth}
    \includegraphics[width=\textwidth]{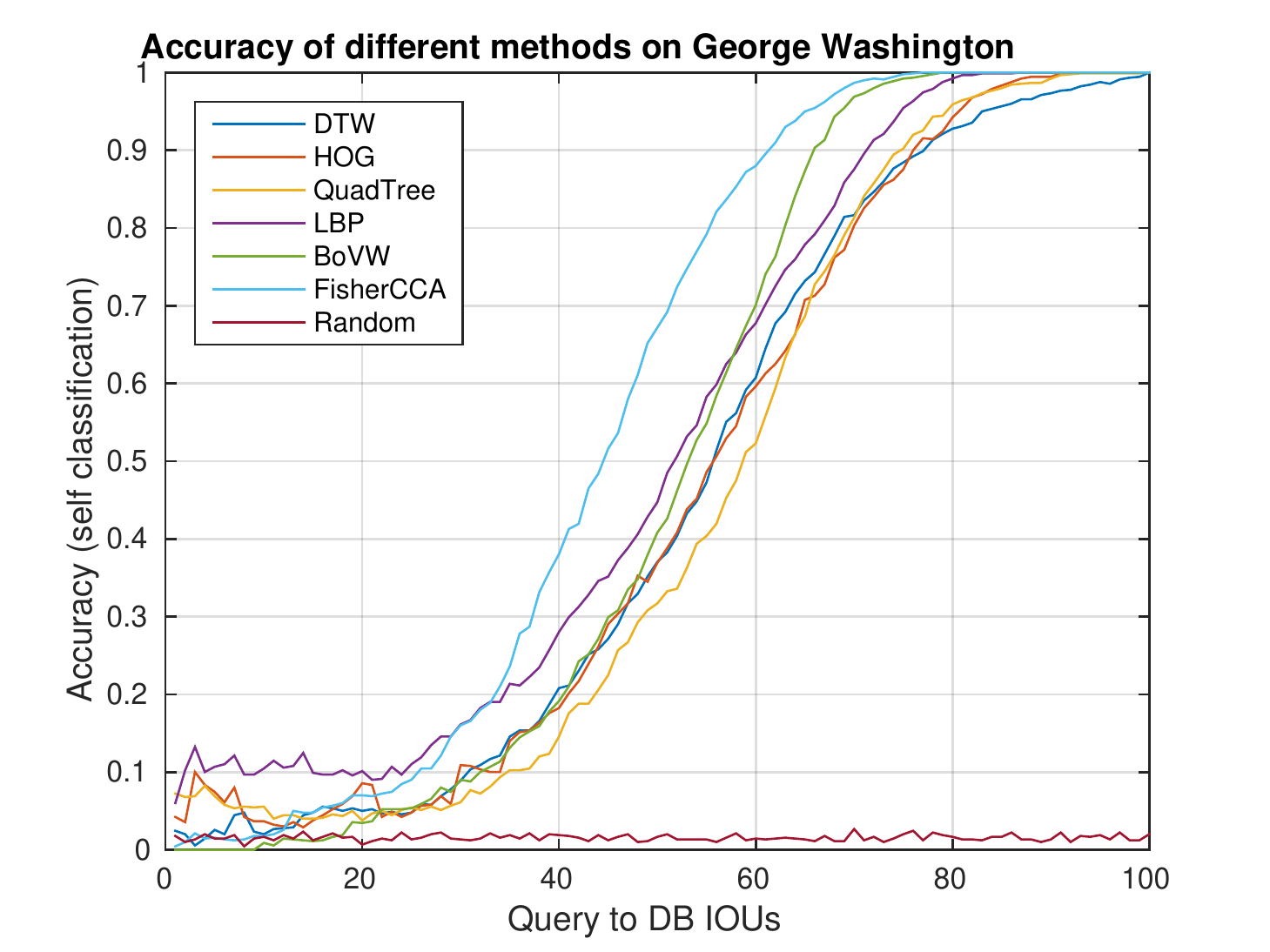}
    \caption{Effect of improper segmentation on self-retrieval for GW.}
	\label{fig:selfclass_GW}
\end{minipage}
\begin{minipage}[b]{0.24\linewidth}
    \includegraphics[width=\textwidth]{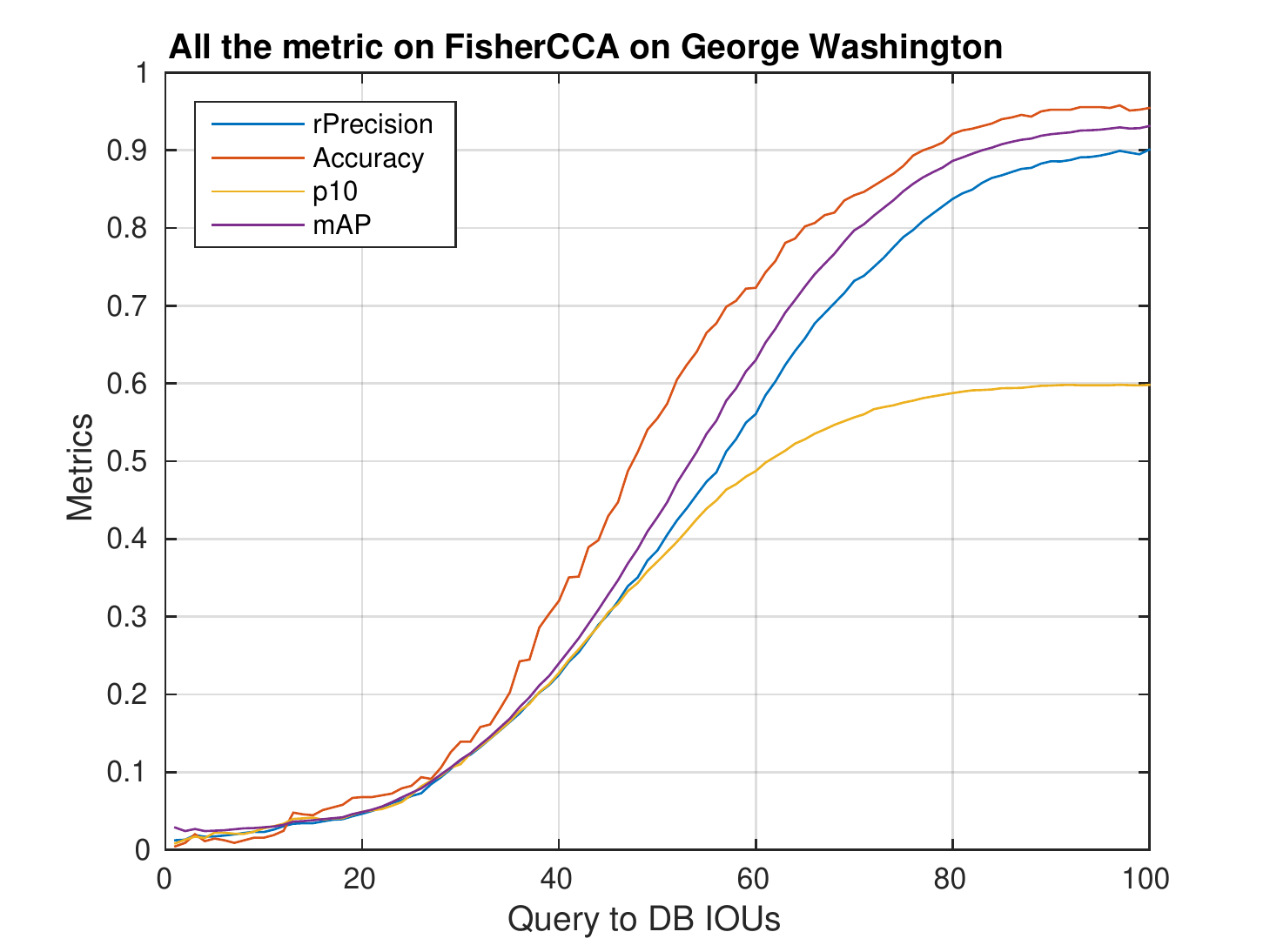}
    \caption{All metrics of Fisher CCA~\cite{almazan2014segmentation} on distorted segmentations.}
   	\label{fig:allMetrics_GW}
\end{minipage}
\end{figure*}

\section{Experimental Analysis}
\label{experiments}
The experiments performed, had as a principal goal to obtain an in depth analysis of the reliance each method has to high quality segmentations.
The principal experiment consists of comparing retrieval of all methods under different levels of distortion on the retrieval database.
A major constraint of designing the experimental procedure was making a fair and informative comparison between supervised learning, unsupervised learning, and learning free methods.

\subsection{Datasets}
As the focus of this work in historical manuscripts, two well established publicly available datasets were used: The George Washington (GW) dataset \cite{fischer2012lexicon}, a single-writer dataset, and the ground-truthed part of the Barcelona Historical Handwritten Marriages Database (BHHMD)\cite{fernandez2014bh2m} which is a multi-writer dataset.
Both datasets were partitioned at the page-level having the first $75\%$ of the pages as train-set and the last $25\%$ of pages as designated as a test-set.
Any words occurring a single time are stemmed when calculating retrieval metrics.
In table~\ref{tbl:ds} the specific word counts for the employed datasets can be seen.
When the dataset is not specified for any measurement, it is performed on GW.

\begin{table}[h]
\label{tbl:ds}
\small
\centering
\caption{Employed Datasets}
\label{tbl:dataset}
\begin{tabular}{p{.14\columnwidth}p{.11\columnwidth}|p{.11\columnwidth}p{.11\columnwidth}p{.11\columnwidth}p{.11\columnwidth}}
\hline
Dataset & Partition & Page\# & Word\#  & Unique Word\# & Stemmed word\# \\ \hline
\multirow{2}{*}{GW\cite{fischer2012lexicon}} & Train & 15 & 3696 & 967 & 265 \\ 
 & Test & 5 & 1164 & 431  & 563 \\ \hline
\multirow{2}{*}{BHHMD\cite{fernandez2014bh2m}} & Train & 30 & 9879 & 1387 & 779 \\ 
 & Test & 10 & 3051 & 607 & 367 \\ \hline
\end{tabular}
\end{table}

\subsection{Metrics}
The quality of word segmentation is quantified as the Intersection over Union (IoU) of the two-point bounding boxes of the proposed word and the word in the Ground-truth.
The definition of IoU for bonding boxes is given in~Eq\ref{eq:iou}.
\begin{equation}
\label{eq:iou}
\centering
\begin{split}
w_I=min(R_{1},R_{2})-max(L_{1},L_{2}))\\
h_I=(min(B_{1},B_{2})-max(T_{1},T_{2}))\\
IoU=\frac{w_I \times h_I}{w_1 \times h_1 + w_2 \times h_2 - w_I \times h_I}
\end{split}
\end{equation}
Where $R_1,L_1,B_1,T_1$ are the sides of the bounding for the first object,  $w_1,h_1$ its width and height etc. .
In Fig.~\ref{fig:alliou} distortions at different IoU can be seen.

Several options were considered as a performance metric for retrievals all derived from precision.
For every query, all the retrieval dataset is ranked by relevance.
Any retrieved samples that share a case-insensitive transcription with the query sample and considered correct.
Precision at any given index is defined as the percentage of correct retrievals for all samples at lesser or equal index.
All performance metrics are estimated by averaging over each query.
The metrics are:
\begin{itemize}
\item{Accuracy: The percentage of queries who's nearest retrieval contains the same transcription, it can be described as precision at index 1.}
\item{rPrecision: The precision each query gets at the retrieval position where a perfect recall and precision scores are possible.}
\item{Precision @ 10: The precision each query gets for the 10 most relevant samples.}
\item{mAP: The average of precision each query gets at each correct retrieval.}
\item{Self-classification accuracy: It is obtained by allowing a query sample to retrieve its self as well, when there is no distortion, all samples by definition obtain a self-classification rate of $100\%$; it is therefore well suited as a metric that ignores performance allowing a comparison of methods with different performances such as learning-free and supervised learning.}
\end{itemize}
For this paper we consider mAP to be the most informative because it better matches the scenario of searching historical documents.

\subsection{Performance evaluation}
\input{tbl_comparision.tex}
Under the same conditions all methods were evaluated for their performance both in metrics and time.
In table~\ref{tbl:ds} the principal characteristics of all methods can be seen.
The column marked as cross-domain refers to training a method on the GW train-set and testing on the test-set of the BHHMD test-set, this measurement allows to estimate the performance of a supervised learning method on unseen data and allows for a comparison with learning free methods.
In what concerns time cost, retrieval time was estimated on the same system under the same conditions and contains the feature extraction and distance matrix calculation.

\subsection{Quality of Segmetation}
The main motivation behind this experiment is to allow for a reliable estimate of the end-to-end performance that can be obtained given segmented QBE word-spotting method.
Segmentation depends on totally different factors than QBE word-spotting and a historical document can be challenging to different degree on either task.
\subsubsection{Distortion Model}
\begin{figure*}[ht]
\centering
\begin{tabular}{ccccc}
    \includegraphics[width=.18\linewidth]{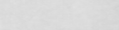} & 	
     \includegraphics[width=.18\linewidth]{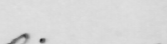} &
     \includegraphics[width=.18\linewidth]{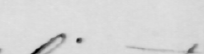} &
     \includegraphics[width=.18\linewidth]{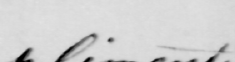} &
     \includegraphics[width=.18\linewidth]{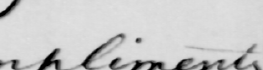} \\
      $10\%$ IoU & $20\%$ IoU & $30\%$ IoU & $40\%$ IoU & $50\%$ IoU\\
      \includegraphics[width=.18\linewidth]{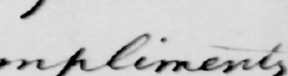} &
      \includegraphics[width=.18\linewidth]{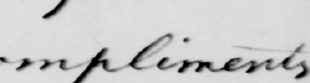} &
      \includegraphics[width=.18\linewidth]{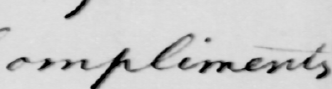} &
      \includegraphics[width=.18\linewidth]{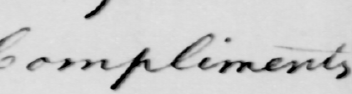} &
      \includegraphics[width=.18\linewidth]{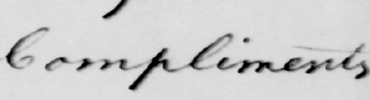} \\
      $60\%$ IoU & $70\%$ IoU & $80\%$ IoU & $90\%$ IoU & $100\%$ IoU\\
  \end{tabular}
\caption{Different IoU of a specific sample}
\label{fig:alliou}
\end{figure*}
The experiment required different degrees of degradation for the retrieval database which in a real word scenario with large data would be automatically segmented while queries selected manually by the user would be well segmented.
A hundred versions of the retrieval database were created, by translating the bounding boxes of the segmentation ground-truth along random directions forcing a specific IoU between the distorted and original sample.
In Fig.~\ref{fig:alliou} the same sample can be observed at different distortions, where $100\%$ IoU means no distortion.
\subsubsection{Distortion Comparison}
The sensitivity to quality of segmentation for every method was computed by applying an identical experimental protocol to all methods.
The train-set was used for any training required by methods including parameter tuning.
For all methods, for every level of distortion, a distance matrix was obtained between all samples in the undistorted test-set and the distorted test-set.
All metrics were computed suppressing the diagonal of the distance matrix and then obtaining the $argsort$ of the distances between each query and all distorted samples.

In Fig.~\ref{fig:mAP_GW} and Fig.~\ref{fig:mAP_BCN} the effect of segmentation on the mAP of each method on GW and BHHMD can be seen.
In Fig.~\ref{fig:allMetrics_GW} the effect of segmentation distortions for all metrics on the best performing method can be seen.
It can be observed that P@10 is diverging from the other three, validating the assumption that it is not as informative for this experiment.
It also observable that accuracy is more unstable than rPrecision and mAP.
In Fig.~\ref{fig:selfclass_GW} self-classification accuracy can be seen.
A qualitative example of a randomly selected query can be seen by all methods in~\ref{fig:retrieval_example}.


\begin{figure}
\includegraphics[width=\columnwidth]{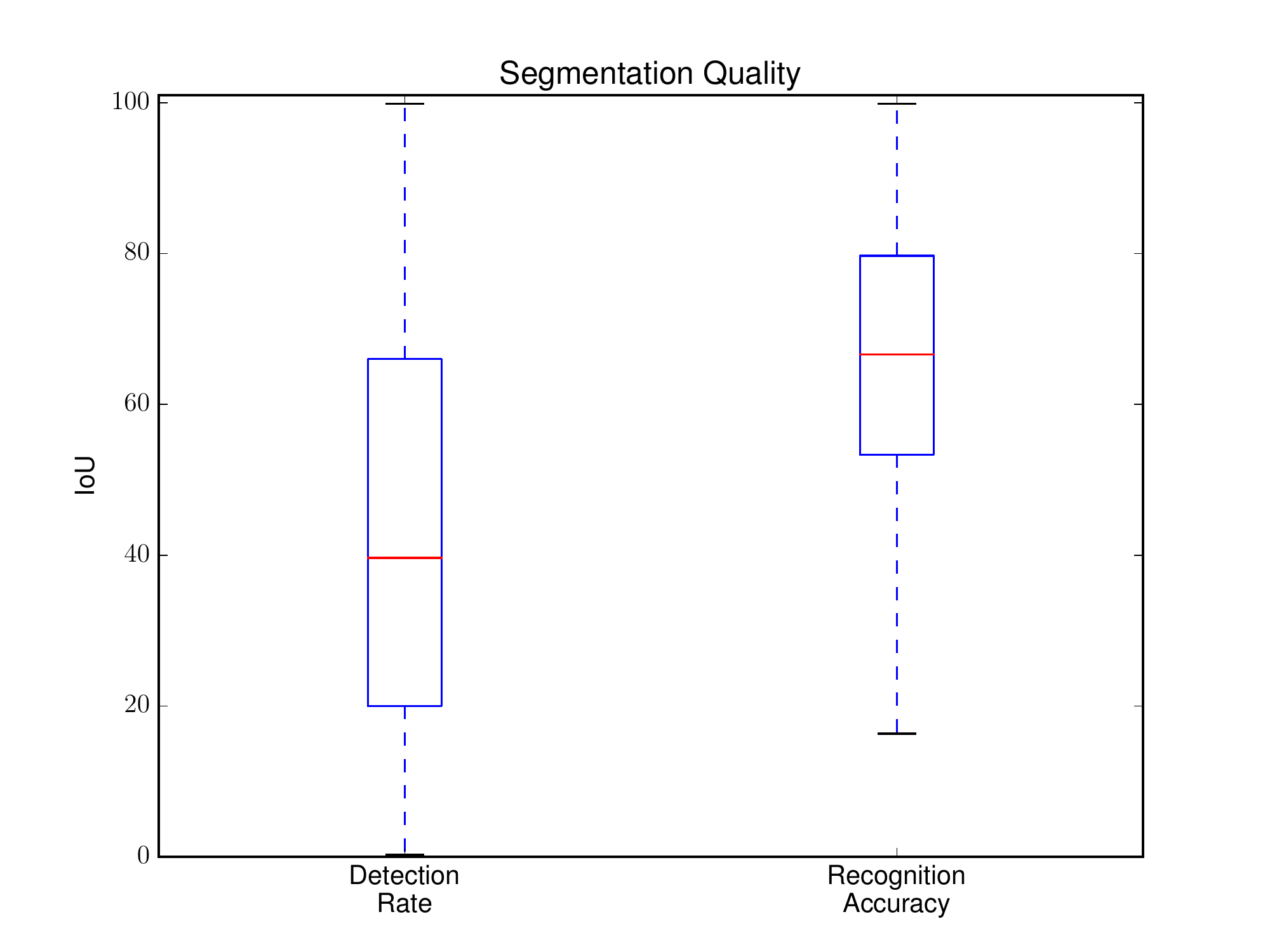}
\caption{Distribution of the IoU obtained by application of
~\cite{manmatha2005scale} 
on GW dataset}
\label{fig:boxplot}
\end{figure}
QoS experiments demonstrated the end-to-end performance of QBE word-spotting can have depending the performance of the word segmentation stage has.
To put the presented curves in context, a binarization-free method for automatic segmentation was used~\cite{manmatha2005scale} executed on the GW dataset, broader range of segmentation methods and a comparative analysis, would go beyond the scope of this paper.
The method was tuned with the GW dataset in mind.
In order to measure the IoU of the proposed words and the actual words, a sparse matrix is created where each row refers to a ground-truth word-box and each column refers to a proposed word-box.
Several measurements can be obtained from this matrix, the most informative with respect to the QoS curves are the column-wise maximum and the row-wise maximum which are related to the Detection Rate and Recognition Accuracy as defined in~\cite{gatos2009segmentation} the only difference being that there is no threshold at $90\%$ and the actual IoU is used as a soft measurement.
In Fig.~\ref{fig:boxplot} the statistics of IoU can be seen.

\begin{figure}[h!]
\begin{minipage}[b]{0.48\linewidth}
    \includegraphics[width=\textwidth]{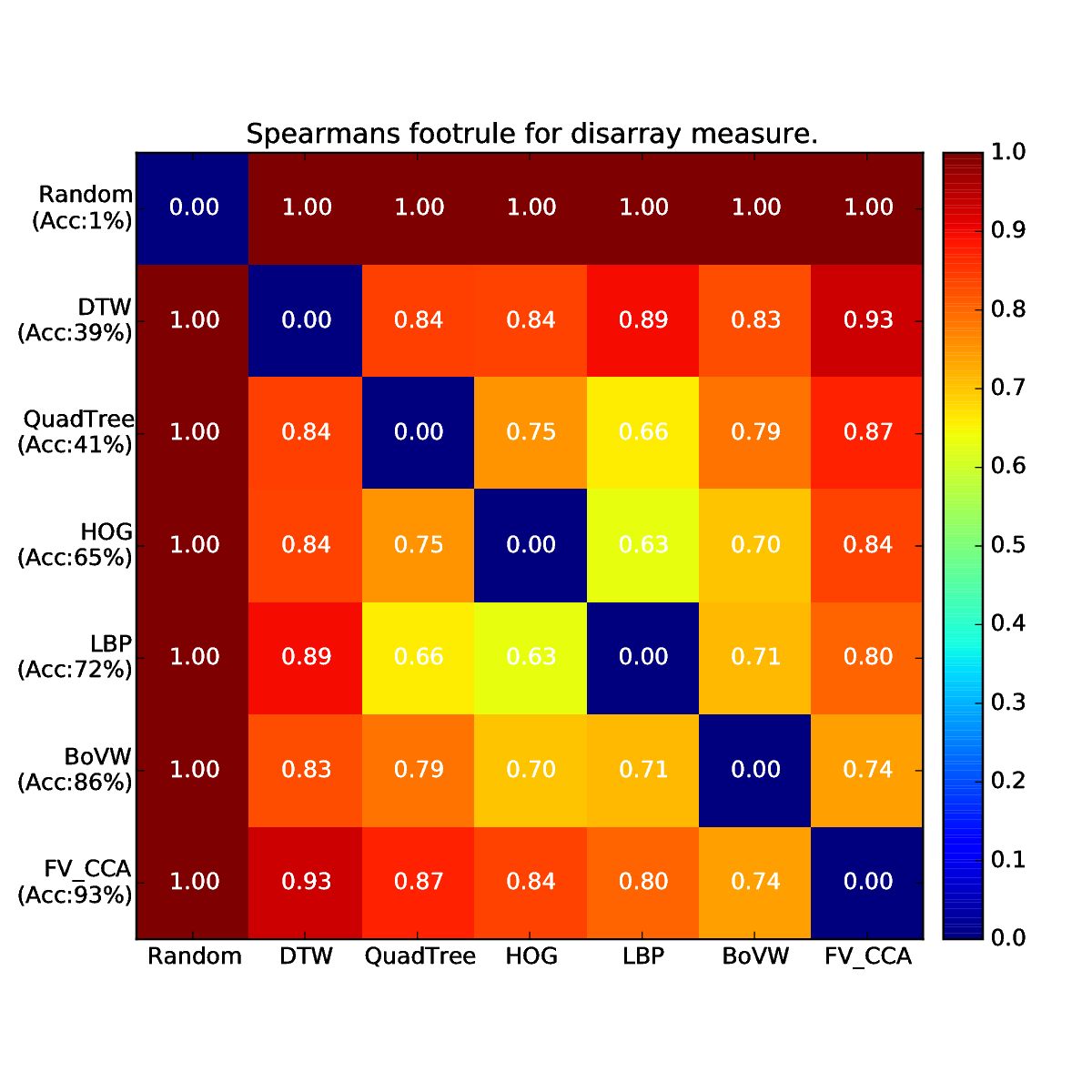}
    \caption{Average Spearman's foot rule between methods.}
    \label{fig:ortho_spearman}
\end{minipage}
\begin{minipage}[b]{0.48\linewidth}
    \includegraphics[width=\textwidth]{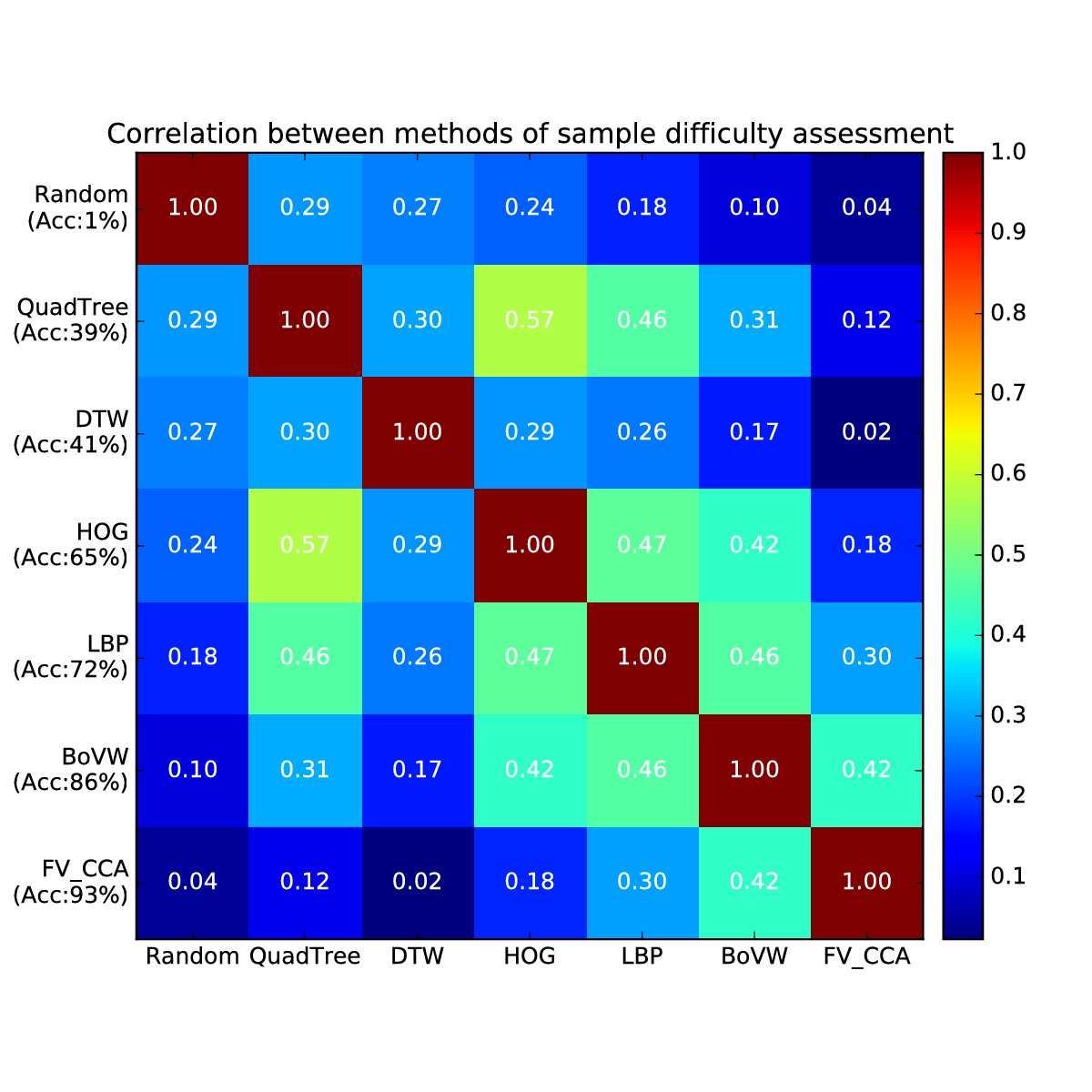}
    \caption{Method agreement in easy and hard samples. Correlation between methods on partitioning the samples in to easy and hard.}
    \label{fig:ortho_ap}
\end{minipage}
\end{figure}


\subsection{Method Independence}
An other part of the experimental analysis of the methods is about gaining insights on the independence of the proposed methods.
The motivation for such an analysis is that it can provide an intuition for the potential that the optimal fusion of methods could provide.
Two measurements are provided as indicators of the independence of the examined methods.
The first is Spearman's footrule~\cite{diaconis1977spearman} measured between the retrievals of two methods.
In Fig.~\ref{fig:ortho_spearman} Spearman's footrule applied on all pairs of methods for the 
GW test-set
can be seen.
Spearman's footrule quantifies the similarity between rankings of samples for every query and therefore it is directly influenced by a methods performance.
A second quantification of the independence of analysed methods is introduced which compensates differences in the methods performance.
For every method, the average precision is calculated for every query.
The query samples are then labelled as 0 if their average precision is lower than the median and as 1 if their average precision is greater than the median average precision.
As an estimate of the dependence, Pearson's correlation is measured on the query-samples labellings.
This measurement in effect quantifies the agreement of two methods on which are the easy samples and which the hard.
In Fig.~\ref{fig:ortho_ap} the correlations on the labelling of the
GW test-set 
can be seen.
In both tables, a random retrieval is added to provide context and scale.

\subsection{Method Fusion}
The weighted linear fusion at the decision level of all analysed methods was evaluated.

\begin{figure}
\label{fig:allfusion}
\includegraphics[width=\columnwidth]{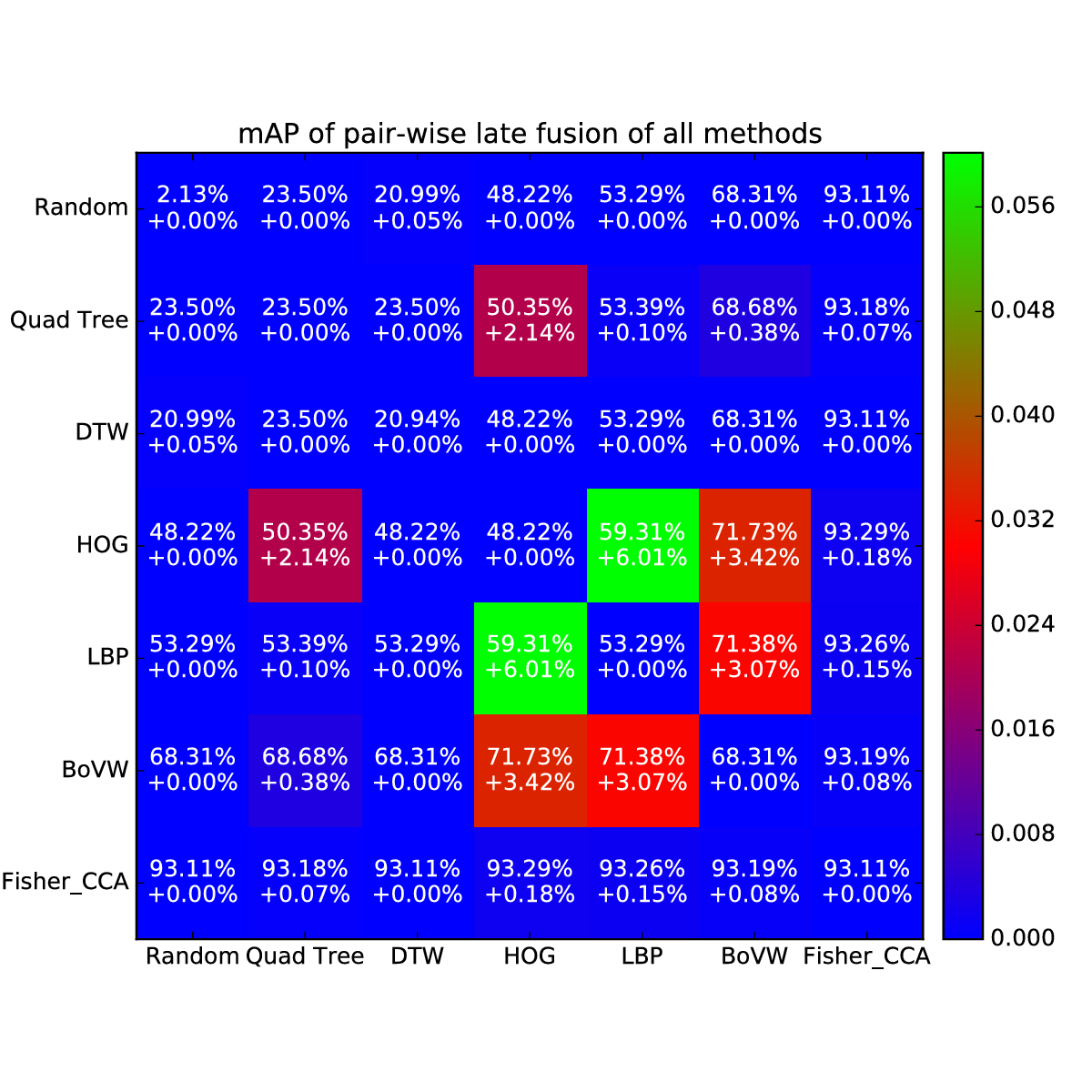}
\caption{mAp increase by weighted linear fusion of methods. The final mAP achieved is marked as a percentage along with the increment in performance due to the fusion.}
\end{figure}

%% file: tbl_comparision.tex
\begin{table*}[!ht]
\small
\centering
\caption{Analysed method performance}
\label{tbl:resultsGW}
\begin{tabular}{r|cccccc}
\hline
\textbf{Method}  & Learning & mAP(GW) & mAP(BCN) & Cross Domain & Retrieval sec. & Train time \\ \hline
Quad-Tree & Standardization & $15.5\%$  & $30.14$  & $15.32$  & $44.41$ & \textbf{0}  \\ 
BoVW \cite{rusinol2011browsing} & Unsupervised & $68.26\%$  & -  & -  & - & hours \\
FisherCCA \cite{almazan2014segmentation} & Supervised & $\textbf{93.11\%}$  & $\textbf{95.40}$  & $\textbf{72.42}$  & $137.63$ & hours  \\ 
DTW \cite{rath2003word} & No & $20.94\%$  & -  & -  & $78095.89$ & \textbf{0} \\ 
HOG pooled Quad-Tree & No & $48.22\%$  & $66.66$  & $66.66$  & $45.34$ & \textbf{0} \\ 
Proposed method (LBP) & No & $54.44\%$  & $70.84$  & $70.84$  & $\textbf{43.17}$ & \textbf{0}\\ \hline
\end{tabular}
\end{table*}

%% file: ch_conclusion.tex
\section{Conclusions and Discussion}
\label{conclusions}
In conclusion it is uncontested that the Fischer-vector method~\cite{almazan2014segmentation} is the best performing method and should be preferred when ground-truth is available and a reasonably performing segmentation is available.
In both datasets the method preserved its remarkably high performance down to an average IoU of $80\%$ after which an almost linear decrease in performance was observed yet it is superior to all other methods down to practically unusable segmentations.
The BoVW~\cite{rusinol2015efficient} method is the best, at least in single writer datasets when no ground-truth is available although it should be pointed out that learning-free methods outperform it when IoU get under $50\%$.
In what concerns learning-free methods the LBP and HOG methods demonstrated comparable performance although LBP was consistently a bit better.
These two methods should be preferred when computation time is important. As demonstrated with the cross-domain measurements they can reach the performance of the Fischer vector method on unseen data which indicates they very well suited for vast and heterogeneous data.
DTW other than mediocre performance stood-out because of its prohibiting time cost.
Even though the matlab implementation could be partially blamed for this, the algorithmic complexity it demonstrates being roughly $O(n^2m)$ where $n$ is the average word width and $m$ is the database size, makes practically unusable in any large scale scenario.

The fusion of methods proved to be a straight forward way of combining methods although improvement could only be obtained in some cases.
The methods that benefited the most, were the HOG and LBP which were quite similar in nature.
It could be hypothesised that a more elaborate fusion in a supervised manner would provide further improvements.

The method independence analysis although interesting, can not be interpreted as a predictor of the improvement achieved by fusion.
One can only speculate on whether the employed metrics were not informative or if a smarter fusion method was needed to harvest that independence.

Focusing on end-to-end performance, the segmentation IoU obtained by~\cite{manmatha2005scale} indicate segmentation is probably the stage that needs improvement, the assumption of high-quality word segmentation being available for historical manuscripts is probably wrong.